\pdfoutput=1

\documentclass[11pt]{article}

\usepackage[final]{acl}

\usepackage{times}
\usepackage{latexsym}

\usepackage[T1]{fontenc}

\usepackage[utf8]{inputenc}

\usepackage{microtype}

\usepackage{inconsolata}

\usepackage{hyperref}
\usepackage{url}
\usepackage{colortbl}
\usepackage{xtab}
\usepackage{longtable}
\usepackage{multirow}
\usepackage{longtable}
\usepackage{booktabs}
\usepackage{amsmath}
\usepackage{amssymb}
\usepackage{verbatim}
\usepackage{xspace}
\usepackage{subfigure}
\usepackage{float}
\usepackage{enumitem}
\usepackage{graphicx}
\usepackage{caption}
\usepackage{wrapfig}
\usepackage{float}
\usepackage[]{todo} 

\definecolor{limegreen}{rgb}{0.5, 1.0, 0.0} 

\def\figref#1{Figure~\ref{fig:#1}}
\def\figlabel#1{\label{fig:#1}\label{p:#1}}

\def\tabref#1{Table~\ref{tab:#1}}

\def\tablabel#1{\label{tab:#1}\label{p:#1}}
\def\eqref#1{Eq.~\ref{eqn:#1}}

\def\Secref#1{Section~\ref{sec:#1}}
\def\Appref#1{Appendix~\ref{sec:#1}}
\def\seclabel#1{\label{sec:#1}\label{p:#1}}
\def\applabel#1{\label{sec:#1}\label{p:#1}}
\def\secref#1{\S\ref{sec:#1}}
\def\appref#1{\S\ref{sec:#1}}

\newcommand{\github}{\raisebox{-1.5pt}{\includegraphics[height=1.05em]{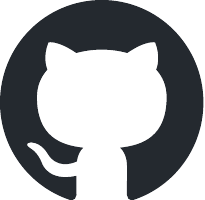}}\xspace}
\newcommand{\hf}{\raisebox{-1.5pt}{\includegraphics[height=1.05em]{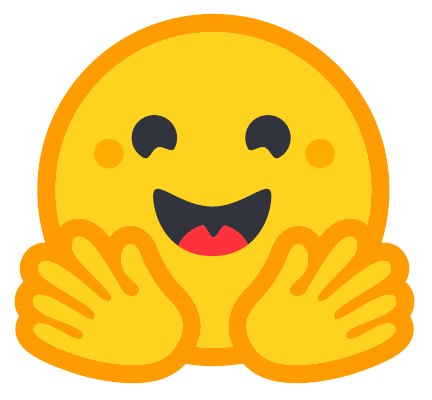}}\xspace}

\def\genericname{\mbox{\textsc{MEXA}}\xspace}

\title{\genericname: Multilingual Evaluation of English-Centric LLMs via Cross-Lingual Alignment}

\author{
    Amir Hossein Kargaran$^{1}$ \quad
    Ali Modarressi$^{1}$ \quad
    Nafiseh Nikeghbal$^{2}$ \\
    \textbf{Jana Diesner}$^{2}$ \quad
    \textbf{François Yvon}$^{3}$ \quad
    \textbf{Hinrich Schütze}$^{1}$ \\
    \\
    $^1$LMU Munich \& Munich Center for Machine Learning \\
    $^2$Technical University of Munich \\
    $^3$Sorbonne Université \& CNRS, ISIR \\
    \texttt{\{amir, ali\}@cis.lmu.de}, \quad
    \texttt{nafiseh.nikeghbal@tum.de}
}

\begin{document}
\maketitle
\begin{abstract}
English-centric large language models (LLMs) often show strong multilingual capabilities. However, their multilingual performance remains unclear and is under-evaluated for many other languages. Most benchmarks for multilinguality focus on classic NLP tasks or cover a minimal number of languages.
We introduce \genericname, a method for assessing the multilingual capabilities of pre-trained English-centric LLMs using parallel sentences, which are available for more languages than existing downstream tasks. \genericname leverages that English-centric LLMs use English as a pivot language in their intermediate layers. \genericname computes the alignment between English and non-English languages using parallel sentences to evaluate the transfer of language understanding from English to other languages. This alignment can be used to estimate model performance in different languages.
We conduct controlled experiments using various parallel datasets (FLORES-200 and Bible), models (Llama family, Gemma family, Mistral, and OLMo), and established downstream tasks (Belebele, m-MMLU, and m-ARC). We explore different methods to compute embeddings in decoder-only models.
Our results show that \genericname, in its default settings, achieves an average Pearson correlation of 0.90 between its predicted scores and actual task performance across languages.
This suggests that \genericname is a reliable method for estimating the multilingual capabilities of English-centric LLMs, providing a clearer understanding of their multilingual potential and the inner workings of LLMs.

\begin{center}
\renewcommand{\arraystretch}{1.2}
\footnotesize
\begin{tabular}{ll}
     \hf \, \textbf{Leaderboard} &\href{https://cis-lmu-mexa.hf.space}{\path{cis-lmu-mexa.hf.space}} \\
     \github \, \textbf{Code} &\href{https://github.com/cisnlp/MEXA}{\path{github.com/cisnlp/MEXA}} \\
\end{tabular}
\end{center}
\end{abstract}

\section{Introduction}\seclabel{introduction}

Most state-of-the-art autoregressive large language models (LLMs) are ``English-centric'', including closed-source models such as GPT-4~\citep{achiam2023gpt}, open-weight models such as Llama 3~\citep{meta2024llama3-1}; and open-source models such as OLMo~\citep{groeneveld2024olmo}. English-centric refers to the majority of the pre-training data for these models being in English~\citep{zhong2024beyond, kew-etal-2024-turning}. Even models labeled as heavily multilingual, such as BLOOM~\citep{workshop2023bloom176}, have their major pre-training data in English~\citep{laurenccon2022bigscience}.
 
Except for open-source models, where pre-training data is available and language distribution is transparent, there is still confusion about the language capabilities and coverage of other LLMs.
Primarily, the focus in evaluating LLMs has been on developing benchmarks to assess their performance in English. Most benchmarks in multilingual settings consist of classical monolingual NLP tasks such as sequence labeling~\citep{ahuja-etal-2023-mega, lai-etal-2023-chatgpt}, automatic translation of popular English benchmarks such as MMLU~\citep{hendrycks2021mmlu} into a limited number of languages~\citep{lai-etal-2023-okapi, openai_mmmlu}, or the creation of language-specific benchmarks~\citep{ghahroodi2024persianmmlullm, koto2024arabicmmlu, son2024koreanmmlu, yuksel2024turkishmmlu, li2024cmmlu}.

Most LLMs are English-centric, either by choice or due to the availability of abundant data sources in English. For these models to be effective in other languages, the other languages must align with the dominant language, i.e., English. Given such alignment, English could act as a \textit{``rising tide that raises all ships,''} meaning that improvements in English performance could benefit other languages, especially in tasks such as reasoning~\citep{zhu2024question}.
Contrarily, if a language does not align well with English, an English-centric LLM may not provide \textit{meaningful coverage} for that language. Indeed, \citet{wendler2024llamas} have found that for Llama 2~\citep{touvron2023llama2}, an English-centric LLM, English could be seen as a kind of ``pivot'' language, enabling to solve complex semantic tasks in a foreign language through a detour into English. More precisely, they show that Llama 2 was able to decode semantically correct next tokens in the middle layers, assigning higher probabilities to the English tokens than to the foreign version, which is only selected in the upper layers. \citet{zhao2024large} present a hypothesis regarding the middle layers of English-centric LLMs, suggesting that these models use English as a means of reasoning while incorporating multilingual knowledge. Based on their analysis, the number of language-specific neurons in the middle layers decreases within the self-attention mechanism but remains consistent across the layers of the feed-forward structure when processing multilingual queries.

In this paper, we introduce \genericname, a method to estimate the actual multilingual coverage of English-centric LLMs. It builds on the observation that these models semantically use English as a pivot language in their middle layers, by measuring how well embeddings of non-English sentences align with their English counterparts.

We verify the \genericname estimation of language coverage for each LLM, using Pearson correlation between estimated and actual scores for various tasks. We use two parallel datasets: FLORES-200 \citep{nllbteam2022} and Bible \citep{mayer-cysouw-2014-creating}; nine LLMs: Llama family, Gemma family, Mistral, and OLMo; and three tasks: Belebele~\citep{bandarkar2024belebele}, m-MMLU, and m-ARC \citep{lai-etal-2023-okapi}. 
Our results show that \genericname achieves a promising average Pearson correlation of 0.90 with established downstream tasks across nine models and two parallel datasets. In our study on the calculation of \genericname scores, we conduct multiple design analyses to examine the impact of token-level pooling for the embeddings (i.e., using the last token versus a weighted average) and layer-level pooling in computing alignment scores. While \genericname demonstrates a high correlation across most setups, we find that a weighted average based on tokens, combined with mean pooling, yields the best results. 
In summary, \genericname offers a scalable way to estimate the multilingual coverage of English-centric LLMs via alignment with English.

\section{Background and Related Work}

We discuss distribution of pre-training data in LLMs, and multilingual evaluation benchmarks in Appendices~\ref{sec:dist-train-llm} and \ref{sec:related-benchmarks}, and focus on cross-lingual alignment here.
Research on cross-lingual alignment either aims to uncover the underlying mechanisms of alignment and assesses its impact on models and downstream tasks, or attempts to enhance model performance by enforcing alignment before, during, or after the pre-training phase. Most of these papers have focused on encoder-only models, such as XLM-R~\citep{conneau-etal-2020-unsupervised} and mBERT~\citep{devlin-etal-2019-bert}, among others~\citep{hammerl-etal-2024-understanding}. In this work, we focus on decoder-only models.

\textbf{Understanding Alignment.}
\citet{ye2023language} show that English-centric models such as Llama~1 \citep{touvron2023llama1} possess multilingual transfer abilities (after fine-tuning on one source language, they can be applied to other languages) and may even surpass the multilingual transfer abilities of multilingual pre-trained models such as BLOOM \citep{workshop2023bloom176}.
\citet{schafer2024imbalance} find that GPT-2-style decoder-only models show strong cross-lingual generalization when trained on an imbalanced mix of languages. However, when trained on a balanced language set, they do not observe increased performance compared to monolingual settings. \citet{wendler2024llamas} perform single-token analysis to demonstrate that English-centered LLMs, such as Llama 2, use English semantically as an internal latent language in the middle layers when handling multilingual queries. \citet{zhong2024beyond} extend this analysis to multiple tokens, also showing that an LLM dominated by both English and Japanese uses both languages as internal latent languages.
\citet{zhao2024large} explore how LLMs handle multilingualism. They hypothesize that LLMs initially interpret the query and convert multilingual inputs into English for task-solving. In the middle layers, the models rely on English with self-attention mechanisms for reasoning, while employing multilingual knowledge through feed-forward structures. In the final layers, LLMs generate responses consistent with the original query language. \citet{li2024quantifying} and \citet{li2024exploring} are even more closely related to ours.
\citet{li2024quantifying} uses absolute cosine similarity values between last token embeddings derived from parallel sentences with English to predict the ranking of language performance across various models. However, as we discuss in \Secref{mexa}, relying solely on absolute cosine values can be misleading, and as shown in \Secref{res-cosine}, absolute cosine values are less correlated with downstream tasks than \genericname score. \citet{li2024exploring} uses English probing tasks and their automatic translations to construct a multilingual evaluation. While they compare embedding similarity scores between high- and low-resource languages with corresponding evaluation results, similar to \citet{li2024quantifying}, they do not assess whether these correlations hold across other downstream tasks. In \Secref{results}, we demonstrate that \genericname scores align closely with a broad range of downstream tasks.

\textbf{Boosting Alignment.} 
The idea to enforce alignment in encoder-only models using parallel sentences dates back to \citet{conneau-lample-2019-xlm}, and has been explored under various guises, e.g., using mixed-language sentences and/or bilingual dictionaries \citep{huang-etal-2019-unicoder,conneau-etal-2020-emerging,cao-etal-2020-multilingual, kulshreshtha-etal-2020-cross, efimov-etal-2023-theimpact, zhang-etal-2023-veco}. Recently,
\citet{li2024prealign} improved multilingual alignment by initializing the decoder-only models to generate similar representations of aligned words using contrastive learning and preserving this alignment using a code-switching strategy during pre-training. \citet{liu2024towards} propose a data allocation technique to select a core subset of languages for fine-tuning, better aligning the multilingual capabilities of decoder-only LLMs and making them more truthful in their responses. \citet{li-etal-2024-improving-context} propose aligning internal sentence representations across different languages using multilingual contrastive learning and aligning outputs by following cross-lingual instructions in the target language for decoder-only models.

\section{\genericname}\seclabel{mexa}

We now describe the \genericname method for computing the alignment score of language \(L_1\) with a pivot language \(L_2\), given the language model \(m\). In this paper, we use the term \textit{cross-lingual alignment}, or simply \textit{alignment} to refer to the semantic similarity of multilingual embeddings across languages. \( L_2 \), for English-centric LLMs and in this paper, is English. To assess alignment, we use parallel sentences in two languages, \( L_1 \) and \( L_2 \). The goal of semantic similarity is to ensure that parallel sentences have sufficiently high similarity, reflecting alignment between the two languages. However, considering only the absolute cosine similarity value as the alignment score does not guarantee proper alignment. For some languages, even non-parallel sentences exhibit similarity scores comparable to those of parallel sentences (see \secref{res-cosine}).
This is largely due to the anisotropy problem observed in transformer models, which can lead to so-called hubness issues, making it difficult to distinguish between similar and dissimilar embeddings \citep{ethayarajh-2019-contextual}, especially in multilingual models \citep{haemmerl-etal-2023-exploring,rajaee-pilehvar-2022-isotropy}.
However, a direct comparative analysis of the cosine similarity between parallel and non-parallel sentence pairs across languages can help overcome these issues. Instead of using the absolute cosine similarity value for alignment, we assign binary values (1 or 0) based on whether a criterion for semantic similarity is satisfied. Our criterion imposes that (a) parallel sentences should have high cosine similarity, and (b) non-parallel pairs should also have low similarity values, ensuring the similarity is not random or biased. Specifically, if the cosine similarity for a pair of parallel sentences is higher than for any non-parallel sentences, we assign a value of 1 for this pair; otherwise, a value of 0. This approach sidesteps the hubness problem since the absolute cosine similarity values themselves are not directly used. 

To compute \genericname, we first apply the cosine similarity function to the pairs of embeddings of parallel sentences in languages \( L_1 \) and \( L_2 \). In \Secref{sent-embd}, we describe how embeddings can be computed for each layer \( l \) of the autoregressive language model \( m \). We generate a square matrix \( C(L_1, L_2, m, l) \) representing cosine similarities of embeddings at the output of layer \( l \) for all parallel sentences in languages \( L_1 \) and \( L_2 \).
We denote \( c_{ij} (l) \) the element in the \( i \)-th row and \( j \)-th column of \( C(L_1, L_2, m, l) \). It represents the cosine similarity between the \( i \)-th sentence of \( L_1 \) and the \( j \)-th sentence of \( L_2 \) at layer \( l \) of language model \( m \). The diagonal elements of \( C \), denoted \( c_{ii} (l) \), represent the cosine similarity between parallel sentence pairs from
\( L_1 \) and \( L_2 \). We define the \genericname alignment score \(\mu \big(C(L_1, L_2, m, l)\big) \) as follows:
\[
\begin{split}
\frac{1}{n} \sum_{i=1}^n \mathbf{1} \bigg( c_{ii} (l) > 
&\quad \max_{j \in \{1, \dots, n\} \setminus \{i\}} \{c_{ij} (l), c_{ji} (l)\} \bigg),
\end{split}
\]

where \( n \) is the number of diagonal elements (i.e., the dimension of the matrix), and \( \mathbf{1}(\cdot) \) is the indicator function, which equals 1 if its argument condition evaluates to true and 0 otherwise. This alignment score measures how often \( c_{ii} (l) \) 
is the maximum value in both its row and column. 
The \genericname\ alignment score can alternatively be understood as 
a measure of sentence retrieval performance \citep{pmlr-v119-hu20b, liu2024transliterations, hammerl-etal-2024-understanding}, with the metric of P@1 applied with queries in language \( L_1 \) and answers in \( L_2 \), and vice versa. We discuss other ways to calculate semantic similarity between languages in \Appref{other-semantic-sim}.

\textbf{Layer-wise Pooling.} The \genericname alignment score \(\mu \big(C(L_1, L_2, m, l)\big)\) is computed for language \(L_1\) respect to pivot language \(L_2\) for each layer \(l\) of the language model \(m\). To compute a single \genericname alignment score given the language model \(m\) and \(L_1\), \(L_2\), we use mean and max pooling strategies over multiple layers. 

\subsection{Sentence Embeddings}\seclabel{sent-embd}

We focus on autoregressive language models that use a decoder-only architecture. In this architecture, attention is not bidirectional; instead, it takes the form of causal attention (left-to-right). In bidirectional attention, each token has access to every other token in the sequence. However, in causal attention, the embedding of a token at position \(t\) is only influenced by the embedding of preceding tokens at positions \( 0, 1, \ldots, t - 1 \). Therefore, simple averaging values biases the embeddings towards sentence-initial words. Instead, we consider alternative methods: using only the last token and weighted averaging. We use and compare both methods to acquire the sentence embeddings needed for \genericname.

A standard way to compute a sentence embedding uses only the last token of that sentence. \citet{jiang2023scaling} show that using the last token in the format of a prompt template for a sentence $s$, such as 'This sentence: \{s\} means in one word:', can be effective. Inspired by this, \citet{li-li-2024-aoe} used the prompt 'Summarize sentence \{s\} in one word:' to obtain the last token embedding as the sentence-level text embedding. However, not all models are instruction-tuned; some earlier works, such as \citet{neelakantan2022text, wang-etal-2024-improving-text, ma2024fine}, use the last token without any prompt. Since the models studied in this paper are only pre-trained and use multiple languages in the input, we decided to use the last token method without any preceding instruction.
An alternative is weighted averaging, which relies on the intuition that using only the last token might not represent the entire sentence, as the influence of earlier tokens may have diminished. This implies that the tokens at the end of the sentence should contribute more to the overall embedding than those at the beginning. Another motivation for using weighted averaging is that sentence-final tokens are influenced by preceding tokens and contain more context, while the representation of sentence-initial tokens has significantly less contextual representation. To address this, \citet{muennighoff2022sgpt} proposes to assign weights to each token based on its position. Thus, the sentence embedding of layer \( l \) using position-weighted averaging is:
\[
e_l = \sum_{t=1}^{T} w_t h_{lt} \quad \textrm{with} \quad w_t = \frac{t}{\sum_{k=1}^{T} k},
\]

where \( T \) is the number of tokens in the given sentence, \( h_{lt} \) is the embedding of the \( t \)-th token at layer \( l \), and \( e_l \) is the sentence embedding at layer \( l \).

\begin{figure}[t]
    \centering
    \includegraphics[width=0.7\linewidth]{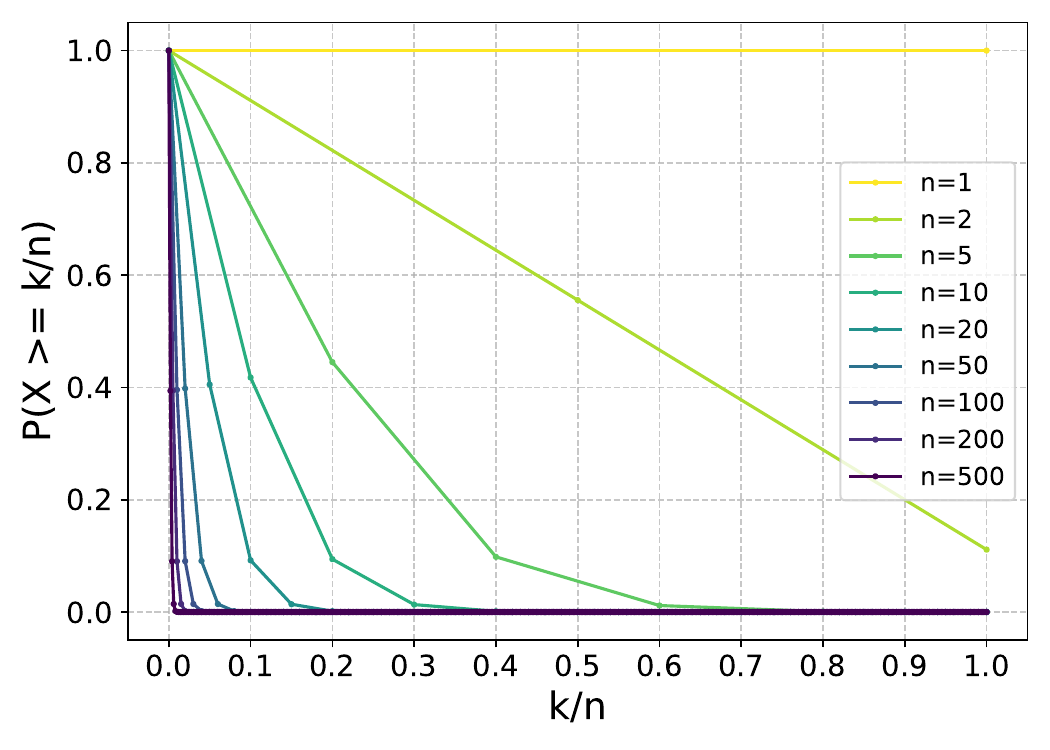}
    \caption{The probability that at least \( k \) out of \( n \) diagonal elements in an \( n \times n \) random matrix are the maximum elements in their respective rows and columns.}
    \figlabel{binomial-probability}
\end{figure}

\subsection{Robustness of \genericname}\seclabel{robustness}

We show that the \genericname alignment score ($\mu(.)$) is very robust, and the odds of this score randomly achieving a high value are very slim. Recall that \(\mu \big(C(L_1, L_2, m, l)\big) \) measures the fraction of diagonal elements
in matrix \( C(L_1, L_2, m, l) \) that have the maximum value in their respective rows and columns. If this condition is met \(k\) times out of \(n\) diagonal elements, then \(\mu \big(C(L_1, L_2, m, l)\big) \) is \(\frac{k}{n}\).
In an \(n \times n\) random matrix, the probability of a diagonal element being the maximum in its row and column (a total of \(2n - 1\) elements) is \(p = \frac{1}{2n - 1}\). The probability that at least \( k \) out of \( n \) independent variables are satisfied, given that the diagonal element is the maximum in its row and column, can be computed using the binomial distribution:
\[
P(X \geq \frac{k}{n}) = 1 - \sum_{i=0}^{k-1} \binom{n}{i} p^i (1 - p)^{n - i}
\]
In \figref{binomial-probability}, we plot \(P(X \geq \frac{k}{n})\). This plot illustrates that, given a sufficient number of parallel sentences (\(n\)), the probability of achieving a high score by chance is very low. For example, with \(n=100\), the chance of obtaining \genericname alignment score larger than $0.05$ (\(k=5\)) from a \(100 \times 100\) random matrix is \(P(X \geq 0.05) = 0.00016\).

\section{Experiments}

We conduct experiments using various multi-parallel datasets (FLORES-200 and the Bible), models (Llama family, Gemma family, Mistral, and OLMo), and existing benchmarks/tasks (Belebele, m-MMLU, m-ARC). Our objective is to assess how well the \genericname alignment score from various parallel datasets correlates with the different benchmarks/tasks for different models.

\subsection{Parallel Data}\seclabel{exp-parallel}

We calculate the \genericname score using the parallel datasets of FLORES-200 \citep{nllbteam2022} and the Bible \citep{mayer-cysouw-2014-creating}. While there are other high-quality parallel datasets, such as NTREX-128~\citep{federmann-etal-2022-ntrex}, IN22 \citep{galaindictrans2}, OPUS-100~\citep{zhang-etal-2020-improving}, Europarl~\citep{koehn-2005-europarl}, OpenSubtitles~\citep{lison-tiedemann-2016-opensubtitles2016}, TED2020~\citep{reimers-gurevych-2020-making}, and Tatoeba~\citep{tatoeba}, we chose FLORES-200 due to its high quality and support for a wide range of languages, and the Bible dataset was selected for its extensive language coverage.

FLORES-200 is a parallel corpus with English sentences from Wikimedia translated into 204 language-script pairs, verified by humans. It includes 997 dev, 1012 dev-test, and 992 test sentences. As the test set isn’t public, we use the dev-test set as our test corpus, following prior work. For faster computation, we consider only the first 100 sentences from each language. As shown in \Secref{robustness}, this is sufficient to ensure MEXA's robustness, as the odds of the \genericname score randomly achieving a high value with 100 sentences are very slim. This choice also enables scaling to more languages, many of which lack enough parallel samples.

The Parallel Bible~\citep{mayer-cysouw-2014-creating} covers a very large number of languages. From this resource, we managed to create a subcorpus, a super parallel dataset of the Bible, with 1,401 language-script labels, each containing 103 sentences (i.e., Bible verses).\footnote{\href{https://huggingface.co/datasets/cis-lmu/sPBC}{\path{hf.co/datasets/cis-lmu/sPBC}}} This corpus includes many low-resource languages, many of which are not covered by existing language technologies~\citep{joshi-etal-2020-state}, and \genericname can be adopted since only parallel data is needed. We use all 103 sentences from each language.

\subsection{Models}

For our experiments, we select models with around 7B parameters, which are considered a base size in the LLM community. The state-of-the-art open-weight models in this range, as measured by performance on English-based tasks such as MMLU~\citep{helm}, include Llama 1, 2, 3, and 3.1~\citep{touvron2023llama1, touvron2023llama2, meta2024llama3-1}, Gemma 1 and 2~\citep{team2024gemma, google2024gemma2}, Mistral 0.3~\citep{jiang2023mistral}, and the open-source model OLMo 1.7~\citep{groeneveld2024olmo}.
We also select a larger model, Llama 3.1 70B, to show that our findings hold even when scaling up further. To apply \genericname, we need to access model weights to compute input sentence embeddings for each layer. We use three popular open-weight model families: Llama, Gemma, and Mistral. As a less multilingual version of state-of-the-art LLMs, we include OLMo, which is trained on a more English-centric corpus of Dolma~\citep{soldaini2024dolma}.

\begin{table*}[t]
\centering
\scriptsize
\resizebox{1.0\linewidth}{!}{
\begin{tabular}{llcccccccccc}
\toprule
& &  Gemma 2 & Gemma 1 & Llama 3.1 & Llama 3.1 & Llama 3 & Llama 2 & Llama 1 & Mistral 0.3 & OLMo 1.7 & AVG \\
& & 9B & 7B & 70B & 8B & 8B & 7B & 7B & 7B & 7B \\
\midrule
\multirow{3}{*}{{$\text{Task}_{\text{\scalebox{.99}{\{eng\}}}}$}}
& Belebele & \underline{0.9178} & 0.8467 & \textbf{0.9456} & 0.8767 & 0.8689 & 0.4822 & 0.4156 & 0.8389 & 0.7711 & 0.7737  \\
& m-MMLU & \underline{0.6998} & 0.6138 & \textbf{0.7700} & 0.6315 & 0.6294 & 0.4523 & 0.3569 & 0.5988 & 0.5210 & 0.5859 \\
& m-ARC  & \underline{0.6775} & 0.5870 & \textbf{0.7014} & 0.5794 & 0.5836 & 0.5128 & 0.5000 & 0.5862 & 0.4872 & 0.5795 \\
\midrule
\multirow{3}{*}{{$\text{Task}_{\text{\,\scalebox{.99}{L$\setminus$\{eng\}} }}$}}
& Belebele  & \underline{0.7093} & 0.5633 & \textbf{0.7684} & 0.5705 & 0.5533 & 0.3028 & 0.2755 & 0.4457 & 0.3627 & 0.5057\\
& m-MMLU  & \underline{0.5582} & 0.4734 & \textbf{0.6384} & 0.4720 & 0.4664 & 0.3260 & 0.2807 & 0.4207 & 0.3390 & 0.4416\\
& m-ARC  & \underline{0.4779} & 0.4220 & \textbf{0.5054} & 0.3941 & 0.3892 & 0.3174 & 0.2970 & 0.3662 & 0.2731 & 0.3825\\
\midrule
\multirow{2}{*}{{FLORES}}
& $\mu_{\text{Mean}}$  & \textbf{0.5088} & 0.3815 & \underline{0.4110} & 0.3963 & 0.3939 & 0.0866 & 0.1946 & 0.2642 & 0.0413 & 0.2976\\
& $\mu_{\text{Max}}$  & \underline{0.7194} & 0.5872 & \textbf{0.7725} & 0.6538 & 0.6520 & 0.2464 & 0.3579 & 0.4716 &  0.1965 & 0.5175 \\
\multirow{2}{*}{{Bible}}
& $\mu_{\text{Mean}}$  & \textbf{0.3568} & 0.2152 & \underline{0.3169} & 0.2103 & 0.2026 & 0.1246 & 0.0908 & 0.1198 & 0.0121 & 0.1832 \\
& $\mu_{\text{Max}}$  & \underline{0.6076} & 0.4021 & \textbf{0.6599} & 0.4212 & 0.4190 & 0.2724 & 0.2357 & 0.2606 & 0.0319 & 0.3678 \\
\bottomrule
\end{tabular}
}
\caption{$\mu_{\text{pooling}}$ shows \genericname scores for each pooling strategy using token-weighted embeddings. Results cover English-only tasks, non-English tasks (Belebele: 116 languages, m-MMLU: 33, m-ARC: 31), and \genericname scores from FLORES (116) and Bible (101). Top scores are in \textbf{bold}, second-best are \underline{underlined}.
}
\tablabel{task}
\end{table*}

\subsection{Benchmarks}

Among the existing evaluation benchmarks in \tabref{all-benchmarks} from \Appref{related-benchmarks}, we chose Belebele \citep{bandarkar2024belebele}, m-ARC \citep{lai-etal-2023-okapi}, and m-MMLU \citep{lai-etal-2023-okapi}, which support the highest number of high-, medium-, and low-resource languages and are directly related to natural understanding tasks, which is the primary focus of this paper.

We use the entire test set for each of these benchmarks (\secref{benchmark-details} for details) for more details) to evaluate the models, except in one case. For Llama 3.1 70B, we use the first 500 questions of m-MMLU instead of the whole set due to resource constraints. Since the selected LLMs used in our experiment are not instruction-tuned, we use 5-shot in-context learning with the lm-evaluation-harness framework, employing log-likelihood-based multiple-choice scoring. Other settings, such as prompt templates, are configured according to the framework's default~\citep{eval-harness, biderman2024lessons}.

\subsection{Evaluation Measures}

We calculate the Pearson correlation coefficient to assess the strength of the correlation between \genericname and downstream performance on our evaluation benchmarks. This coefficient is a statistical measure of the strength and direction of the linear relationship between two variables. A high value would indicate that \genericname provides a reliable assessment of multilingual capabilities in English-centric LLMs.

\section{Results \seclabel{results}}

\tabref{task} presents the downstream performance of the selected models across three benchmarks, along with \genericname scores from two parallel datasets. Notably, among models with parameter sizes ranging from 7 to 9 billion, both Gemma 2 and Llama 3.1 outperform the other LLMs in terms of non-English downstream performance and \genericname scores. The Llama 3.1 and Llama 3 models exhibit similar alignment and downstream task performance, and both represent substantial advancements compared to Llama 2.
Moreover, results for the Llama 3.1-70B model indicate that scaling can significantly enhance alignment when compared to its smaller version. Interestingly, while Mistral achieves comparable results to Gemma 1 on English benchmarks, it demonstrates inferior alignment, which likely accounts for its reduced performance on non-English tasks.
Furthermore, the Llama 2 model achieves higher \genericname scores than OLMo, indicating better alignment. However, due to Llama 2's weaker performance on English tasks, it fails to transfer this alignment effectively, leading to comparable non-English performance between Llama 2 and OLMo. This observation is further explored in \Secref{downstream-estimation}, where we normalize the expected performance based on the pivot language, namely English.

\subsection{\genericname Correlation Analysis}\seclabel{results-correlation}

\begin{table}[t]
\centering
\scriptsize
\resizebox{0.85\linewidth}{!}{
\begin{tabular}{llc}
\toprule
& & Avg. across models \\
\midrule
\multirow{12}{*}{\rotatebox{90}{FLORES}}
\multirow{6}{*}{\rotatebox{90}{\scalebox{.65}{weighted average}}}
& $\mu_{\text{Mean}} \leftrightarrow$ Belebele & 0.8994 \\
& $\mu_{\text{Max}} \leftrightarrow$ Belebele & \underline{0.9098} \\
& $\mu_{\text{Mean}} \leftrightarrow$ m-MMLU   & \underline{0.9513} \\
& $\mu_{\text{Max}} \leftrightarrow$ m-MMLU   & 0.9188 \\
& $\mu_{\text{Mean}} \leftrightarrow$ m-ARC    & \textbf{0.9393} \\
& $\mu_{\text{Max}} \leftrightarrow$ m-ARC    & 0.8856 \\
\cmidrule{2-3}
\multirow{12}{*}{\rotatebox{90}{\phantom{FLORES}}}
\multirow{6}{*}{\rotatebox{90}{\scalebox{.65}{last token}}}
& $\mu_{\text{Mean}} \leftrightarrow$ Belebele & \textbf{0.9168} \\
& $\mu_{\text{Max}} \leftrightarrow$ Belebele & 0.9058 \\
& $\mu_{\text{Mean}} \leftrightarrow$ m-MMLU   & \textbf{0.9545} \\
& $\mu_{\text{Max}} \leftrightarrow$ m-MMLU   & 0.9134 \\
& $\mu_{\text{Mean}} \leftrightarrow$ m-ARC    & \underline{0.9195} \\
& $\mu_{\text{Max}} \leftrightarrow$ m-ARC    & 0.8685 \\
\midrule
\midrule
\multirow{12}{*}{\rotatebox{90}{Bible}}
\multirow{6}{*}{\rotatebox{90}{\scalebox{.65}{weighted average}}}
& $\mu_{\text{Mean}} \leftrightarrow$ Belebele & \underline{0.8496} \\
& $\mu_{\text{Max}} \leftrightarrow$ Belebele & \textbf{0.8811} \\
& $\mu_{\text{Mean}} \leftrightarrow$ m-MMLU   & \textbf{0.8823} \\
& $\mu_{\text{Max}} \leftrightarrow$ m-MMLU   & \underline{0.8210} \\
& $\mu_{\text{Mean}} \leftrightarrow$ m-ARC    & \textbf{0.9018} \\
& $\mu_{\text{Max}} \leftrightarrow$ m-ARC    & \underline{0.8354} \\
\cmidrule{2-3}
\multirow{12}{*}{\rotatebox{90}{\phantom{Bible}}}
\multirow{6}{*}{\rotatebox{90}{\scalebox{.65}{last token}}}
& $\mu_{\text{Mean}} \leftrightarrow$ Belebele & 0.8147 \\
& $\mu_{\text{Max}} \leftrightarrow$ Belebele & 0.8070 \\
& $\mu_{\text{Mean}} \leftrightarrow$ m-MMLU   & 0.7572 \\
& $\mu_{\text{Max}} \leftrightarrow$ m-MMLU   & 0.6998 \\
& $\mu_{\text{Mean}} \leftrightarrow$ m-ARC    & 0.7469 \\
& $\mu_{\text{Max}} \leftrightarrow$ m-ARC    & 0.6885 \\
\bottomrule
\end{tabular}
}
\caption{Pearson correlation between \genericname scores and performance on Belebele, m-MMLU, and m-ARC, averaged across models. Results use two embedding aggregation methods: weighted average and last-token. Best scores per dataset and benchmark are in \textbf{bold}, second-best are \underline{underlined}.
}
\tablabel{pearson_avg}
\end{table}

We compute sentence embeddings for the selected models using two methods: weighted average based on token positions and last token (see~\secref{sent-embd}).
We apply mean and max pooling on the \genericname alignment scores across all model layers to derive a single score for each language.  In \tabref{pearson_avg} (refer to \tabref{pearson} for the detailed table), we report the correlation between the \genericname scores (computed using both mean- and max-pooling, for the two embedding methods) and task performances. Across all settings, the best overall result (higher correlation) is achieved when embeddings are computed using the \textbf{weighted average}, with \textbf{mean pooling} as the pooling method. We adopt this configuration as the default setting for \genericname.

\textbf{FLORES vs Bible.} In the default setting, the average Pearson correlation coefficient for the FLORES parallel dataset across different tasks is 0.9300, and while for the Bible parallel dataset, it is 0.8779. The reason the Bible scores are generally lower than FLORES is that FLORES data is cleaner and more aligned with modern, standardized texts, whereas the Bible data is older and more specialized. For example, for some languages, the orthography of Bible texts no longer matches today's orthography. In the Bible, Arabic often includes diacritics, which are typically omitted in modern writing and tasks, making the text less familiar to models trained on contemporary data. Additionally, although the Bible dataset has been made parallel, sentence alignment can still be inconsistent due to translation nuances. In contrast, FLORES is carefully curated to ensure high-quality, sentence-level parallelism across languages for machine translation tasks.

\textbf{Weighted Average vs. Last Token Embeddings.} The use of last token embeddings shows promisingly high correlations with the FLORES parallel data; however, for the Bible dataset, the correlation is low in some cases. We believe this may stem from the high occurrence of Bible sentences (especially in English), which leads models to memorize these phrases. Using the WIMBD toolkit \citep{elazar2024whats}, we found that, on average, there are 92 times more documents in Dolma 1.7 containing exact Bible sentences than those in FLORES.
Consequently, when using Bible examples, the last token is biased towards predicting the specific memorized next token rather than incorporating context-related signals. Therefore, one should consider the hazard of memorized data when using last token embeddings. The weighted-average method, which takes into account the influence of multiple tokens, can mitigate the impact of a poor embedding for the last token and enable the model to capture useful information from the other tokens more robustly.

\textbf{Max Pooling vs. Mean Pooling.} In our comparison of mean pooling and max pooling on the Belebele benchmark, we found that mean pooling underestimates low-resource languages (resulting in more \genericname scores near 0), while max pooling correlates better with the Belebele benchmark. This can be explained by the fact that Belebele is an easier task among the three evaluated, allowing even low-resource languages to achieve good scores. Conversely, based on our experiment with m-ARC, max pooling tends to overestimate low-resource languages, making mean pooling more aligned with m-ARC. This can be attributed to m-ARC being the most challenging task among the three, where even medium-resource languages do not achieve high scores. Changing the pooling method from mean to max can be considered when dealing with different levels of understanding.

\subsection{Downstream Performance Estimation} \seclabel{downstream-estimation}

\begin{figure}[t]
    \centering
    \begin{minipage}[b]{0.23\textwidth}
        \centering
        \includegraphics[width=\textwidth]{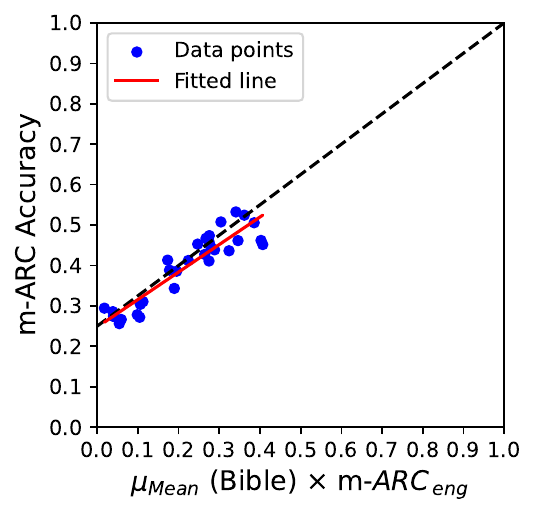} 
    \end{minipage}
    \hfill
    \begin{minipage}[b]{0.23\textwidth}
        \centering
        \includegraphics[width=\textwidth]{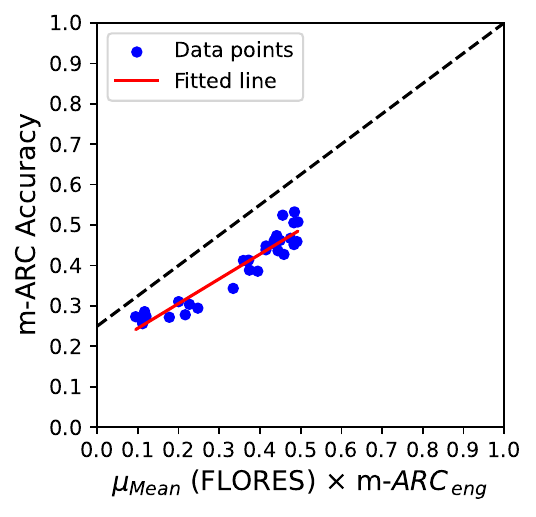}
    \end{minipage}
    \hfill
    \begin{minipage}[b]{0.23\textwidth}
        \centering
        \includegraphics[width=\textwidth]{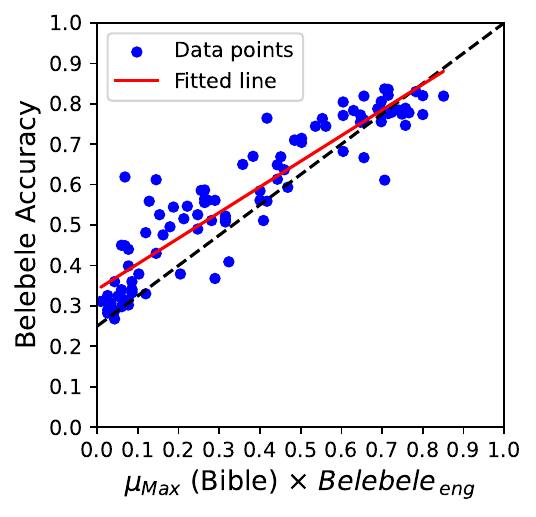}
    \end{minipage}
    \hfill
    \begin{minipage}[b]{0.23\textwidth}
        \centering
        \includegraphics[width=\textwidth]{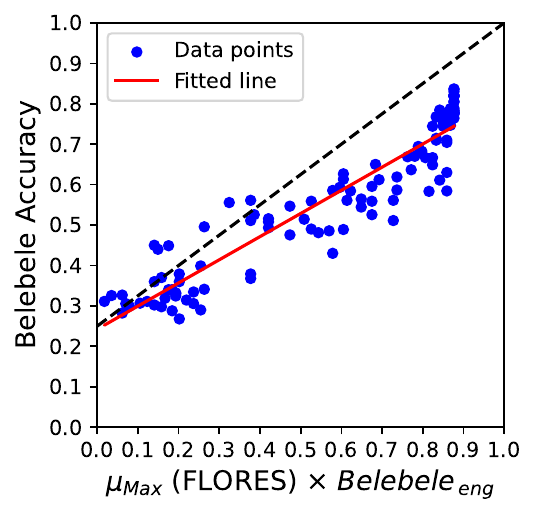}
    \end{minipage}
\caption{Relationship between Llama 3.1-8B \genericname scores from Bible and FLORES, adjusted by English task performance, on Belebele and m-ARC. Each point represents one language.}
    \figlabel{estimation}
\end{figure}

A full score on Pearson correlation (i.e., $\rho=1.0$) indicates that a linear equation perfectly describes the relationship between \genericname and the evaluation benchmarks, with all data points lying on a line. Given the high correlation values shown in \tabref{pearson}, it is reasonable to conclude that we can fit a line that closely approximates this linear relationship.
This line converts the \genericname scores back to downstream task performances. We employed a linear model to predict this line by minimizing the residual sum of squares between the \genericname scores (multiplied by the performance on the English task) and the task performances. We needed to adjust the \genericname scores for this purpose, as the \genericname score for language \(L_1\) indicates how well \(L_1\) is aligned with English but does not reflect the estimated task performance of the model for language \(L_1\). Of course, this does not change the value of the correlation coefficient, as it is unaffected by linear transformations.
The three tasks considered in this paper involve multiple-choice questions with four possible answers for each question, resulting in a chance of being randomly correct of \(\frac{1}{4}\). However, the minimum score for \genericname scores is 0. Thus, the ideal slope for the line would be \(\frac{3}{4}\) with an intercept of \(\frac{1}{4}\) (X-axis: adjusted \genericname scores, Y-axis: task performance). In \figref{estimation}, we plot this relationship for Llama 3.1-8B using the Bible and FLORES parallel datasets for Belebele and m-ARC. We chose max pooling for Belebele and mean pooling for m-ARC, since these pooling methods yield a stronger correlation (see \secref{results-correlation}). The pairs of (slope, intercept) from left to right in the \figref{estimation} are: (0.6804, 0.2477), (0.6103, 0.1838), (0.6340, 0.3408), and (0.5726, 0.2423). With data points from both high-resource and low-resource languages, this line can be calculated; otherwise, the ideal line may be used as a reference. 

\textbf{Language Coverage.} We present the adjusted \genericname score for all languages available in FLORES-200 in \tabref{whole-flores} from \Appref{mexa-whole-flores} for the selected models. The languages are categorized into groups ranging from well-covered to not covered. In \tabref{whole-flores}, we can clearly see that Llama 3.1-70B and Gemma 2-9B show a higher level of multilinguality than other models.

\subsection{\genericname vs Absolute Cosine Similarity}\seclabel{res-cosine}

We compare \genericname with the use of absolute cosine similarities. We used parallel data from FLORES and downstream task data from the Belebele benchmark, focusing on 116 common labels. For each non-English language, we computed the average absolute cosine similarity for parallel sentences with English, and for non-parallel sentences with English. Following the setup by \citet{li2024quantifying}, which uses absolute cosine similarity values to predict language performance and ranking, we computed sentence embeddings using the last-token method and applied mean pooling over layers \{5, 10, 15, 20, 25\}.

To evaluate the correlation of each method with downstream task performance, we report results using the Gemma 1 and Llama 1 7B models. \tabref{cosine_comparison} summarizes these results. For both models, \genericname consistently achieves a higher correlation with downstream performance compared to the absolute cosine similarity of parallel sentences. Moreover, the correlation between cosine similarity for parallel and non-parallel sentences is notably high, suggesting that absolute values may be less discriminative across sentence types.
This discrepancy highlights a limitation of using absolute cosine similarity: for some languages, similarity scores may remain high even for non-parallel sentences. Conversely, a low overall similarity score does not necessarily imply weak alignment, as parallel sentence scores may still significantly exceed non-parallel ones. In contrast, \genericname offers a more robust and comparative measure across languages.

\begin{table}[t]
\centering
\scriptsize
\resizebox{0.85\linewidth}{!}{
\begin{tabular}{lcc}
\toprule
& \textbf{Gemma 1 7B} & \textbf{Llama 1 7B} \\
\midrule
\genericname\ $\leftrightarrow$ Belebele         & 0.9260 & 0.8365 \\
AC-P $\leftrightarrow$ Belebele                  & 0.7651 & 0.6473 \\
AC-P $\leftrightarrow$ AC-NP                     & 0.9232 & 0.9064 \\
\bottomrule
\end{tabular}
}
\caption{
Pearson correlations between alignment metrics and Belebele performance. AC-P denotes absolute cosine similarity of parallel pairs; AC-NP, of non-parallel pairs.}
\tablabel{cosine_comparison}
\end{table}

\subsection{Visualization of Layers}\seclabel{visualization}

In \figref{llama1and3.1-20langs}, we show the results of applying \genericname to 20 pairs of language\_script from the FLORES parallel dataset for Llama 1-7B and Llama 3.1-8B across all 32 layers. We selected these languages from different families, writing systems, and both high- and low-resource categories. The embeddings are computed using a weighted average based on token positions. \figref{llama1and3.1-20langs} shows that high-resource languages (with more prevalence on the web; see \appref{dist-train-llm}) achieve higher alignment scores across different layers, while low-resource languages achieve lower scores. In the initial layers, embeddings are more in-language, resulting in lower alignment scores. As embeddings progress to the mid-layers, they become more aligned with the dominant language of the LLM, i.e., English.

\begin{figure}[t]
    \centering
    \begin{minipage}[b]{0.35\textwidth}
        \centering
        \includegraphics[width=\textwidth]{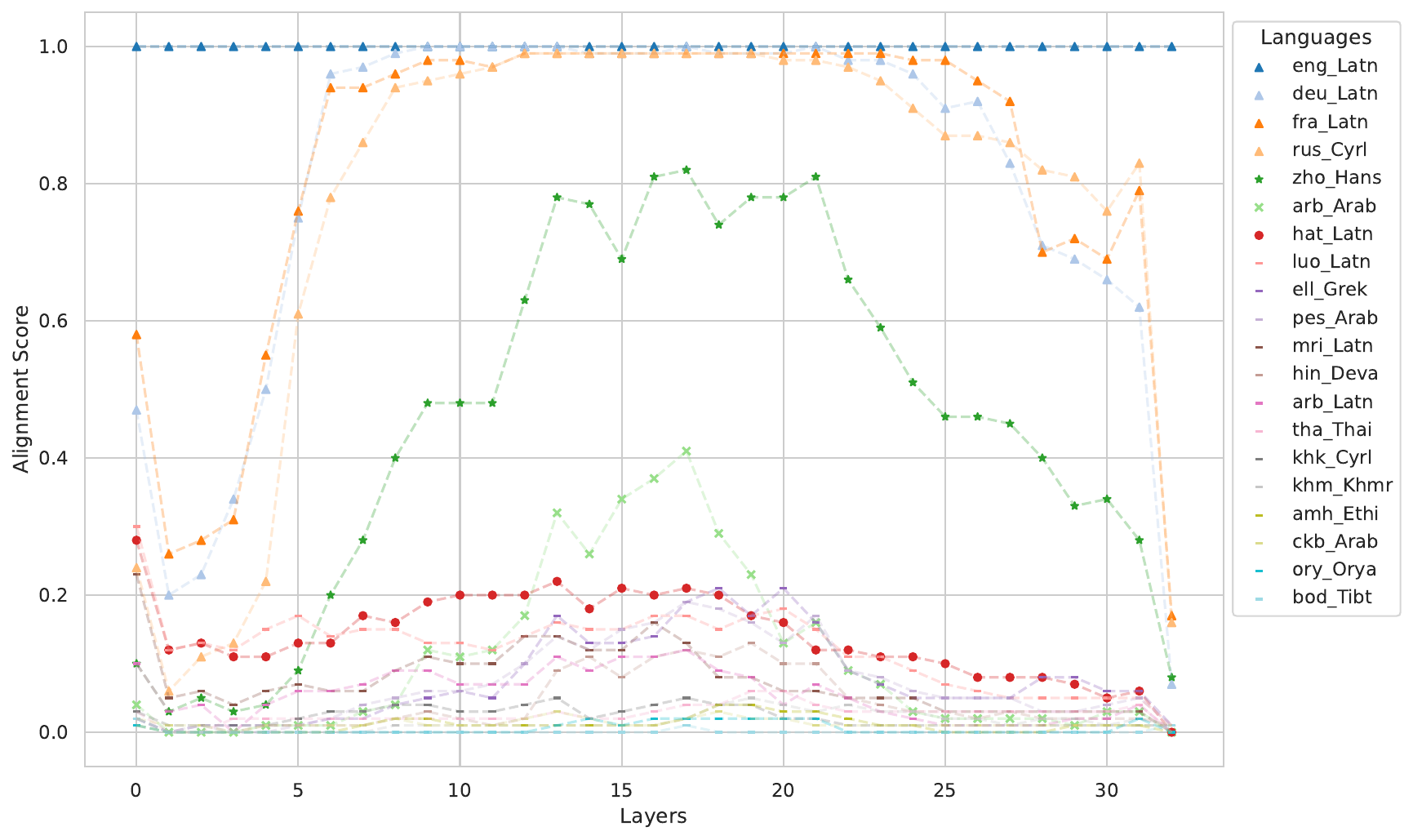} 
        \caption*{Llama 1 7B}
    \end{minipage}
    \hfill
    \begin{minipage}[b]{0.35\textwidth}
        \centering
        \includegraphics[width=\textwidth]{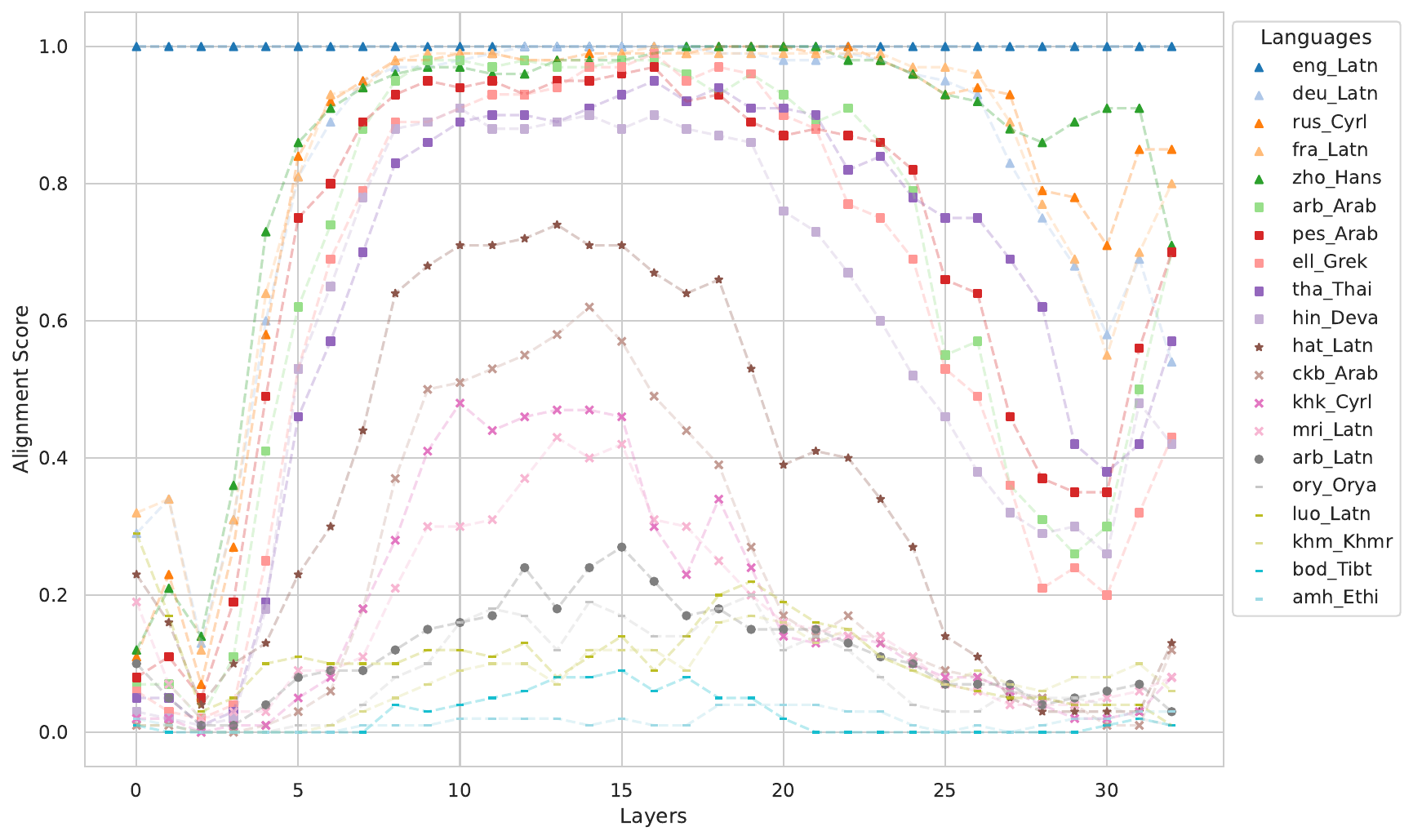}
        \caption*{Llama 3.1 8B}
    \end{minipage}
\caption{Llama 1 vs. Llama 3.1 \genericname alignment score for different languages across all layers. Best performance markers in order: $\triangle$, $\square$, $\star$, $\times$, $\circ$, $\_$}
    \figlabel{llama1and3.1-20langs}
\end{figure}

\begin{figure}[t]
    \centering
    \begin{minipage}[b]{0.22\textwidth}
        \centering
        \includegraphics[width=\textwidth]{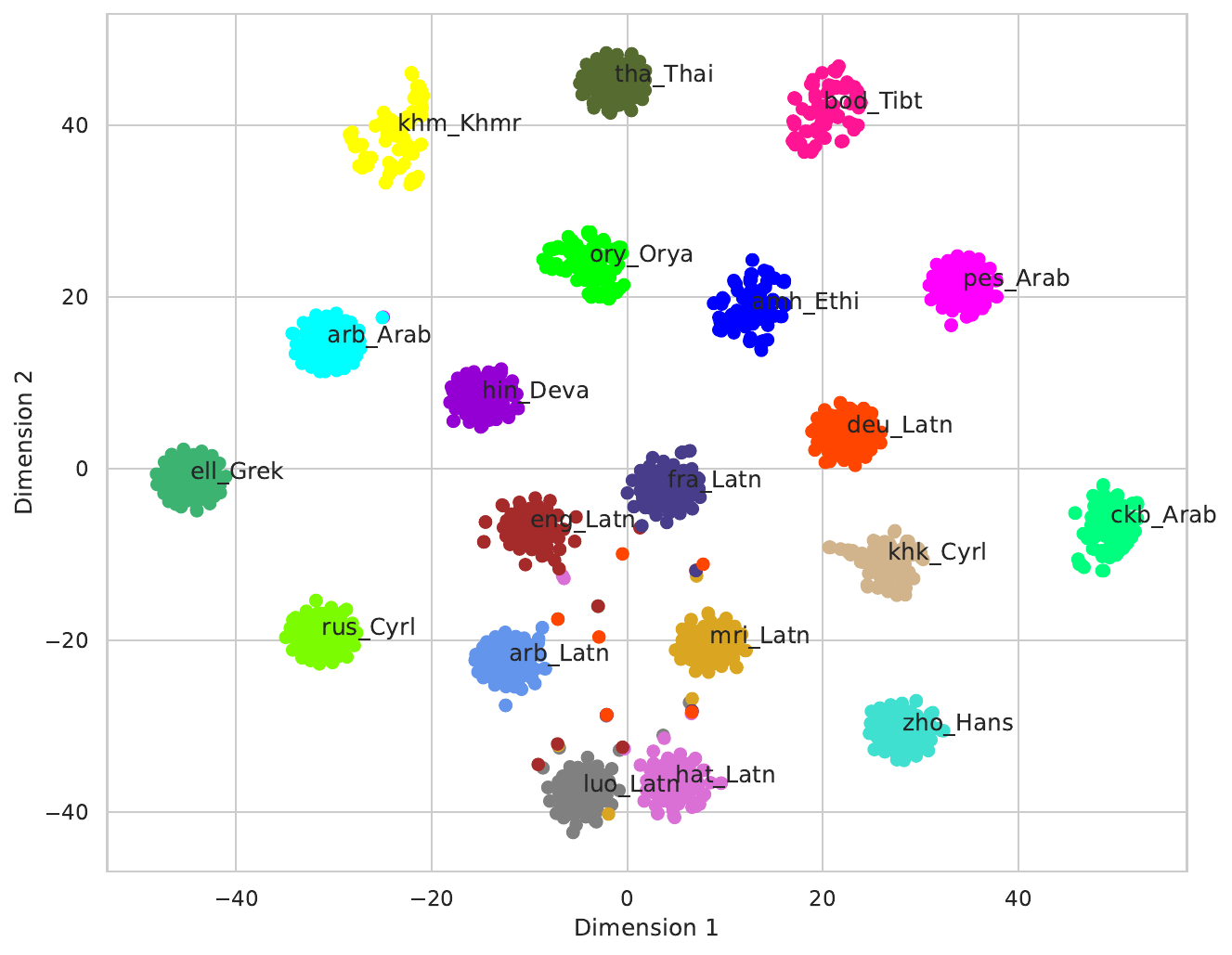} 
        \caption*{Layer 0}
    \end{minipage}
    \hfill
    \begin{minipage}[b]{0.22\textwidth}
    \centering
    \includegraphics[width=\textwidth]{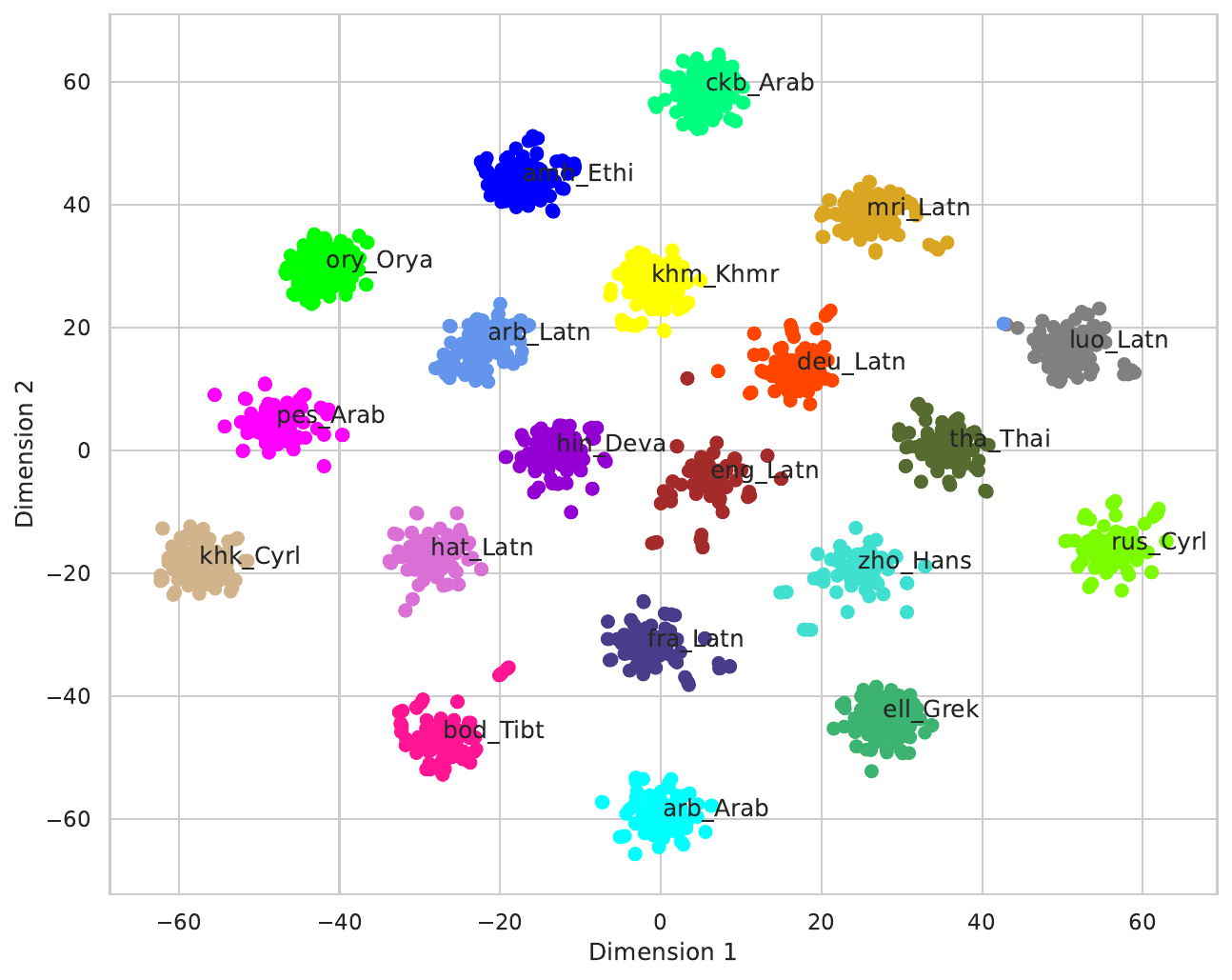}
    \caption*{Layer 32}
    \end{minipage}
    \hfill
    \begin{minipage}[b]{0.3\textwidth}
        \centering
    \includegraphics[width=\textwidth]{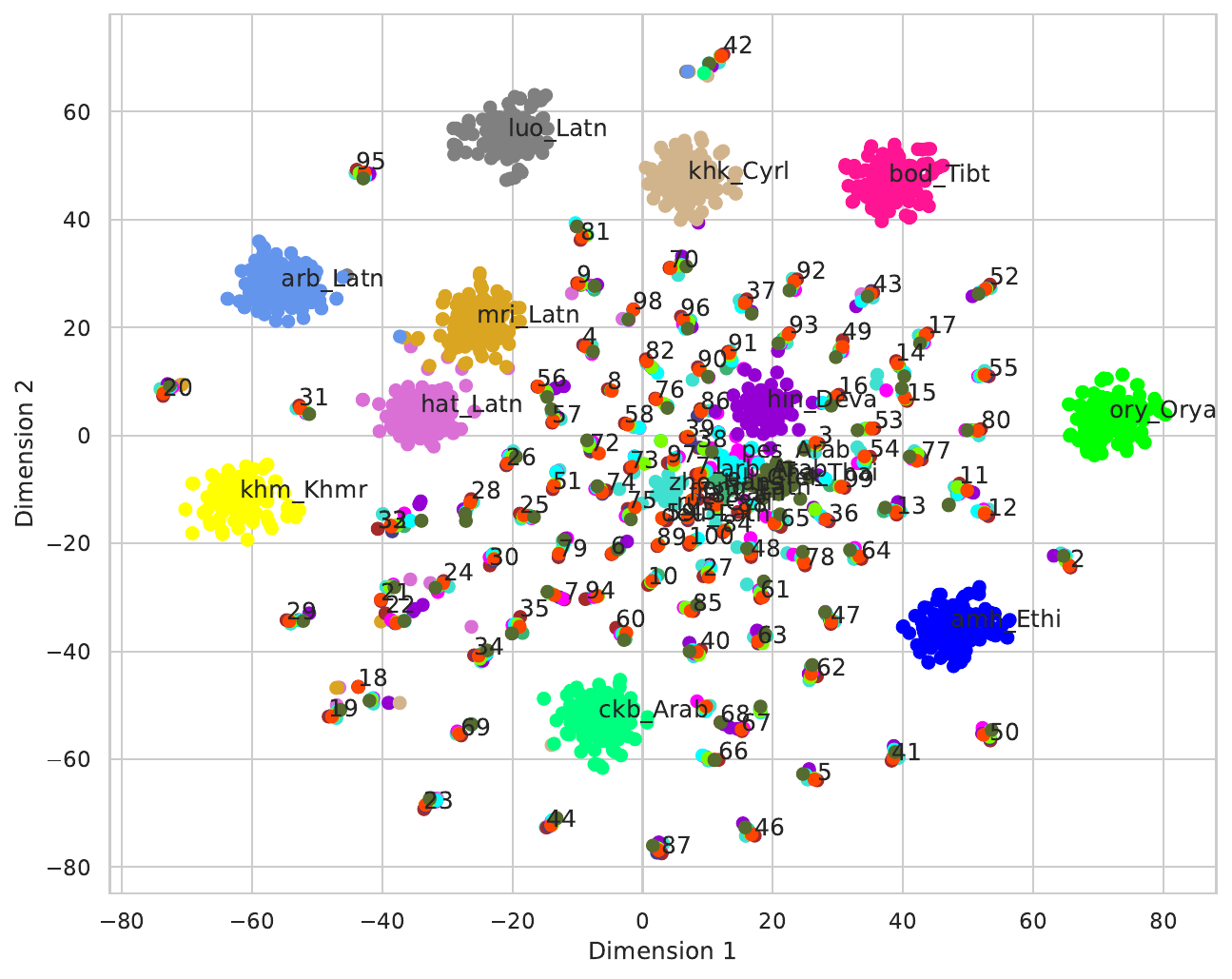}
    \caption*{Layer 13}  

    \end{minipage}
\caption{Llama 3.1 t-SNE plots for 3~different layers. As shown, in the mid-layers, the embeddings become more language-neutral. The numbers shown in the mid-layers are the IDs of English sentences that are scattered.}
    \figlabel{llama3.1set}
\end{figure}

\genericname is comparable between models as long as the same parallel dataset and setting is used to obtain the \genericname scores. \figref{llama1and3.1-20langs} shows that in many languages, particularly high-resource languages, Llama 3.1 achieves a significantly higher alignment score than its predecessor, Llama 1.
Although Llama 3.1 leads to better alignment scores with English for medium and low-resource languages, there is still room for improvement. Comparing Arabic (arb\_Arab) with its romanized version (arb\_Latn), we see that both Llama 1 and Llama 3.1 models perform better in the native script than in the Latin script, even though Llama 1’s tokenizer for Arabic is essentially a character-based tokenizer. In general, for very low-resource languages, those in Latin script tend to have higher alignment scores, likely because the tokenization is more favorable for Latin characters.

In \figref{llama3.1set}, we display the t-SNE \citep{van2008visualizing} plots of the embeddings of \figref{llama1and3.1-20langs} from 3 different layers of Llama 3.1: embedding layer 0, mid-layer 13, and last layer 32. We assign a different color to each language.
For layers 0 and 32, the embeddings are more language-specific, while in the mid-layer, they become more language-neutral. Languages that maintain their language-specific embeddings in the mid-layer are clustered separately and, notably, correspond to the very low-resource languages that receive the lowest alignment scores from \genericname.

\section{Conclusion}

We introduce \genericname, a method for assessing the multilingual capabilities of English-centric large language models (LLMs). \genericname builds on the observation that English-centric LLMs semantically use English as a kind of pivot language in their intermediate layers. \genericname computes the alignment between non-English languages and English using parallel sentences, estimating the transfer of language understanding capabilities from English to other languages through this alignment. This metric can be useful in estimating task performance, provided we know the English performance in the task and the alignment score between languages derived from a parallel dataset. Through different studies with two parallel datasets (FLORES-200 and the Bible), different LLMs including the Llama family, Gemma family, Mistral, and OLMo, and three downstream tasks (Belebele, m-MMLU, and m-ARC), we demonstrated that \genericname provides a reliable estimation of multilingual performance. For \genericname score calculations, multiple design analyses are conducted to explore the impact of token-level pooling for embeddings and layer-level pooling in computing alignment scores. While \genericname shows high correlation across most configurations, a weighted average of tokens combined with mean pooling delivers the best results. The results reveal a promising average value for the Pearson correlation coefficient of 0.90 with established downstream tasks across nine models and two parallel datasets. Overall, \genericname proves to be a valuable method for practitioners aiming to assess the multilingual capabilities of English-centric LLMs, easing future efforts to expand these models to a wider range of underrepresented languages.

\section*{Limitations}
We are aware of four main limitations of our work.

First, The scope of this paper is limited to non-generative tasks. Generation is generally more challenging than understanding, and it is unsurprising that these models for many languages may struggle to generate content in their language. While NLP has advanced toward generative capabilities, a significant portion of evaluation still focuses on non-generative tasks (e.g., Hugging Face Leaderboards)\footnote{\href{http://huggingface.co/spaces/open-llm-leaderboard/open_llm_leaderboard}{\path{hf.co/spaces/open-llm-leaderboard/open_llm_leaderboard}}} due to their convenience in multiple-choice question evaluation and standardized metrics. Assessing generated output remains challenging, even in English benchmarks. For example, model-based approaches (e.g., ``LLM-as-a-judge''~\citep{zheng2023judging}) require an LLM fully competent in the target language---a capability that is both questionable and the focus of our evaluation.

Second, \genericname provides a method of evaluation for open science, and only model weights are needed. Although, developers of closed-source models could use \genericname under the hood and report their multilingual results to provide insight of their model's multilingual capabilities. We target the widely used settings where the LLM follows a decoder-only transformer architecture. For other architectures, as long as we can extract the embedding given a sentence for intermediate layers, \genericname can be calculated.

Third, We present a selection of tasks for multilingual evaluation in \tabref{all-benchmarks}. As shown, for non-generative tasks, only a few benchmarks support a high number of languages, including low-resource ones. Benchmarks limited to around 10 languages, which mostly support high-resource languages, would not support our claims, as \genericname would achieve high results for all of them. Belebele includes the highest number of languages (except sequence labeling tasks), making it an ideal task to evaluate \genericname. Both m-MMLU and m-ARC are the next highest covered languages for non-generative tasks. However, since they are machine-translated tasks, they are not ideal and may bias some results for low-resource languages (or, more accurately, when the machine translation is poor). Yet, these translated versions are representative of the current state of automatic evaluation, as seen in multilingual leaderboards.\footnote{\href{https://huggingface.co/spaces/uonlp/open_multilingual_llm_leaderboard}{\path{hf.co/spaces/uonlp/open_multilingual_llm_leaderboard}}}

Fourth, \genericname provides a rough estimate of the multilingual capabilities of pre-trained English-centric LLMs. Different tasks offer diverse perspectives on the abilities of LLMs, and \genericname cannot replace all of them. Our goal is to highlight the multilingual potential of English-centric LLMs and propose a simple way to evaluate them. We hope this encourages the development of more multilingual LLMs, even though they are likely to contain large shares of English data. Additionally, it is important to note that answers across languages do not always need to be fully aligned~\citep{naous-etal-2024-beer}, and for such cases, language- and culture-specific evaluation benchmarks should be developed.

\section*{Acknowledgments}

This work was funded by Deutsche Forschungsgemeinschaft (project SCHU 2246/14-1).

\bibliography{main}

\appendix

\section{Appendix}

\subsection{Distribution of Pre-training Data in LLMs}\applabel{dist-train-llm}
The distribution of languages in the training data of state-of-the-art LLMs is rarely fully documented. Llama 2~\citep{touvron2023llama2} is a counter-example and its authors have disclosed the language distribution use in pre-training.
Their analysis uses the FastText~\citep{bojanowski-etal-2017-enriching} language identification tool and a threshold  of $0.5$ for the language detection. We reproduce \citet[Table~10]{touvron2023llama2}, which lists 27 languages with percentages greater than 0.005\% in the Llama 2 pre-training data, in \tabref{llama2-dist}. English, with 89.70\%, constitutes the vast majority of the training data. 

All the languages listed in \tabref{llama2-dist} have a presence of more than 0.10\% (top 35 languages) on the web according to the {W3Techs report}~\citep{w3techsstat} or more than 0.15\% (top 36 languages) according to {CommonCrawl (first three snapshots of 2024)}~\citep{commoncrawlstat}.
However, not all of the most prevalent languages on the web appear in \tabref{llama2-dist}. The following 9 languages are missing, most of which use non-Latin writing systems: Turkish (tur\_Latn), Persian (pes\_Arab), Arabic (ara\_Arab), Greek (ell\_Grek), Hebrew (heb\_Hebr), Thai (tha\_Thai), Hindi (hin\_Deva), Slovak (slk\_Latn), and Lithuanian (lit\_Latn). 

The distribution of data in the training of English-centric LLMs is not the same as on the web, but it does have some correlation. The amount of English in LLM pre-training data is significantly larger than for other languages. This is also observable for {GPT-3}~\citep{brown2020language}, where more than 92\%
of the training texts was in English~\citep{gpt3dataset}. The rest of the top languages in the data of such models are mostly high-resource languages, which have the most available data on the web (top 36 languages). However, in some models, this could be adjusted by design, for example, to make writing systems with non-Latin languages less prominent (as seen in Llama 2).
This weakens the correlation between LLMs' pre-training data and the web.

\begin{table}[t]
\centering
\scriptsize
\resizebox{1.0\linewidth}{!}{
\begin{tabular}{llr|llr}
\toprule
Language & Script & \multicolumn{1}{l}{Percent}  & Language & Script & \multicolumn{1}{l}{Percent} \\
\midrule
English (eng)   & Latn    & 89.70\%                     & Ukrainian (ukr)   &  Cyrl   & 0.07\%                      \\
Unknown (unk)   &  -    & 8.38\%                      & Korean (kor)  & Hang        & 0.06\%                      \\
German (deu)  & Latn      & 0.17\%                      & Catalan (cat) & Latn        & 0.04\%                      \\
French (fra)  & Latn      & 0.16\%                      & Serbian (srp)  & Cyrl/Latn       & 0.04\%                      \\
Swedish (swe)  & Latn     & 0.15\%                      & Indonesian (ind)  & Latn    & 0.03\%                      \\
Chinese (zho)  & Hans/Hant     & 0.13\%                      & Czech (ces)    & Latn       & 0.03\%                      \\
Spanish (spa)  & Latn     & 0.13\%                      & Finnish (fin)  & Latn       & 0.03\%                      \\
Russian (rus)  & Cyrl     & 0.13\%                      & Hungarian (hun)  & Latn     & 0.03\%                      \\
Dutch (nld)    & Latn     & 0.12\%                      & Norwegian (nor) & Latn      & 0.03\%                      \\
Italian (ita)  & Latn     & 0.11\%                      & Romanian (ron)  & Latn      & 0.03\%                      \\
Japanese (jpn) & Jpan     & 0.10\%                      & Bulgarian (bul) & Cyrl      & 0.02\%                      \\
Polish (pol)  & Latn      & 0.09\%                      & Danish (dan)  & Latn        & 0.02\%                      \\
Portuguese (por) & Latn   & 0.09\%                      & Slovenian (slv)  & Latn     & 0.01\%                      \\
Vietnamese (vie) & Latn   & 0.08\%                      & Croatian (hrv)  & Latn      & 0.01\%   \\
\bottomrule
\end{tabular}
}
\caption{Language distribution in the pre-training data for Llama 2. The large ``Unknown'' category is partially composed of programming code data. Common scripts are sourced from the GlotScript resource~\citep{kargaran-etal-2024-glotscript}.}
\tablabel{llama2-dist}
\end{table}

\subsection{Multilingual Evaluation Benchmarks}\applabel{related-benchmarks}

Multilingual evaluation methods and the development of benchmarks not only facilitate the assessment of diverse language representations in LLMs but also help in monitoring cross-lingual generalization, to assess the effect of quantization across multiple languages \citep{marchisio2024does}, the development of language-specific models \citep{tejaswi2024exploring}, and the optimization of safety preferences \citep{li2024preference}, among others. In \tabref{all-benchmarks}, we list benchmarks with the largest language coverage. This list includes benchmarks referenced by MEGA \citep{ahuja-etal-2023-mega}, MEGAVERSE~\citep{ahuja-etal-2024-megaverse}, xP3 \citep{muennighoff-etal-2023-crosslingual}, the Aya collection \citep{singh2024aya}, the lm-evaluation-harness framework \citep{eval-harness, biderman2024lessons}, and inter alia.  These datasets comprise a mix of translated datasets, some human-translated or verified by native speakers such as AfriXNLI \citep{adelani2024irokobench} and some relying only on machine translation \citet{lai-etal-2023-okapi}. Additionally, there are datasets created independently for each language, such as XLSum \citep{hasan-etal-2021-xl}, where the data is not parallel and the size of the data varies between languages.
Despite the efforts reflected in \tabref{all-benchmarks}, the community is still lacking highly multilingual benchmarks for tasks such as natural language understanding or text generation. 

\begin{table}[t]
     \centering
     \scriptsize
     \resizebox{1.0\linewidth}{!}{
     \begin{tabular}{lll}
     \toprule
    Dataset&Task& \# L\\
    \midrule
    XNLI~\citep{conneau-etal-2018-xnli}&Natural Language Inference& 15\\
    IndicXNLI~\citep{aggarwal-etal-2022-indicxnli}&Natural Language Inference&11\\
    AfriXNLI~\citep{adelani2024irokobench}&Natural Language Inference&15\\
    m\_HellaSwag~\citep{lai-etal-2023-okapi}&Natural Language Inference&31\\
    PAWS-X~\citep{yang-etal-2019-paws}&Paraphrase Identification&7\\
    XCOPA~\citep{ponti-etal-2020-xcopa}&Commonsense Reasoning& 11\\
    XStoryCloze~\citep{lin-etal-2022-shot}&Commonsense Reasoning& 11 \\
    m-ARC~\citep{lai-etal-2023-okapi}&Common Sense Reasoning&31\\
    TyDiQA~\citep{clark2020tydi}&Question Answering& 11 \\
    MLQA~\citep{lewis2020mlqa}&Question Answering&7\\
    XQuAD~\citep{artetxe2020cross}&Question Answering& 11\\
    IndicQA~\citep{doddapaneni-etal-2023-towards}&Question Answering& 10\\
    AfriQA~\citep{ogundepo-etal-2023-cross}&Question Answering& 10\\
    m\_TruthfulQA~\citep{lai-etal-2023-okapi}&MC General Question Answering&31\\
    UDPOS 2.7~\citep{de-marneffe-etal-2021-universal}&Part of Speech Tagging& 104\\
    WikiANN~\citep{pan2017cross}&Name Entity Recognition&282\\
    XLSum~\citep{hasan-etal-2021-xl}&Summarization&  44 \\
    WikiLingua~\citep{ladhak-etal-2020-wikilingua}&Summarization &18
    \\
    Belebele~\citep{bandarkar2024belebele}&MC Reading Comprehension&115\\  AfriMMLU~\citep{adelani2024irokobench}&MC Knowledge Question Answering&17\\
    m-MMLU~\citep{lai-etal-2023-okapi}&MC Knowledge Question Answering&31\\
    MMMLU~\citep{openai_mmmlu}&MC Knowledge Question Answering&15\\
    M3Exam~\citep{zhang2023m3exam} & MC Multimodal Question Answering & 9 \\
  \bottomrule
     \end{tabular}
     }
     \caption{Multilingual evaluation benchmarks: MC stands for multiple-choice. \# L shows the number of languages supported by each dataset.}
     \tablabel{all-benchmarks}
 \end{table}

\subsection{Semantic Similarity in Multilingual Embeddings}\applabel{other-semantic-sim}

There are other ways to compute similarity between languages, such as Representational Similarity Analysis (RSA)~\citep{chrupala-alishahi-2019-correlating} and Central Kernel Alignment (CKA)~\citep{pmlr-v97-kornblith19a}. RSA involves first computing the cosine similarity for sentence embeddings within each language, then correlating these in-language similarities with those in other languages. CKA, another metric, is adopted by~\citet{conneau-etal-2020-emerging} and~\citet{muller-etal-2021-first}. \citet{conneau-etal-2020-emerging} show that the CKA similarity is highly correlated with sentence retrieval scores for four languages. In this paper, our focus is not on finding different ways to calculate similarity between languages, but on how helpful a properly defined alignment score can be in estimating the multilingual capabilities of LLMs across multiple languages.

\subsection{Benchmark Details}\applabel{benchmark-details}

Belebele is a multiple-choice reading comprehension task designed to assess language models across a range of high-, medium-, and low-resource languages. Each question in the dataset is paired with four possible answers and linked to a brief passage from the FLORES-200  dataset \citep{nllbteam2022}. The human annotation process was carefully curated to generate questions that effectively differentiate between various levels of language comprehension, supported by rigorous quality checks. Belebele supports 122 distinct labels (language-script combinations) corresponding to 115 distinct languages. However, FLORES-200 does not support 5 of these labels, corresponding to Romanized versions of 5~Indic languages. Therefore, we conducted our analysis between the FLORES-200 and Belebele benchmarks on 117 common labels. Additionally, there are 102 common labels between the Bible parallel data and the Belebele benchmark.

Both ARC Challenge~\citep{clark2018think} and MMLU~\citep{hendrycks2021mmlu} are also set up as multiple-choice question-answering tasks, but they focus on different types of knowledge and reasoning skills. ARC Challenge is classified as a common-sense reasoning task, consisting of grade-school level science questions, while MMLU consists of questions across a wide range of fields, including the humanities, social sciences, and more. \citet{lai-etal-2023-okapi} used GPT-3.5-turbo~\citep{openai2022chatgpt} and a translation prompt to translate examples from both datasets and create m-ARC and m-MMLU in 31~languages (excluding English).
Later, m-MMLU was expanded to also include Icelandic (isl\_Latn) and Norwegian (nob\_Latn). The Icelandic portion was translated using the Mideind.is, while Norwegian was generated with DeepL.com.\footnote{\href{https://huggingface.co/datasets/alexandrainst/m_mmlu}{\path{hf.co/datasets/alexandrainst/m_mmlu}}} m-MMLU consists of 277 questions in its training set, 13,258 in the test set and 1,433 in the validation set.
m-ARC consists of 1,116 questions in the training set, 1,169 in the test set, and 298 in the validation set.

\begin{table*}[th]
\centering
\scriptsize
\resizebox{1.0\linewidth}{!}{
\begin{tabular}{llcccccccccc}
\toprule
& &  Gemma 2 & Gemma 1 & Llama 3.1 & Llama 3.1 & Llama 3 & Llama 2 & Llama 1 & Mistral 0.3 & OLMo 1.7 & AVG \\
& & 9B & 7B & 70B & 8B & 8B & 7B & 7B & 7B & 7B \\
\midrule
\multirow{12}{*}{\rotatebox{90}{FLORES}}
\multirow{6}{*}{\rotatebox{90}{\scalebox{.65}{weighted average}}}
& $\rho$ ($\mu_{\text{Mean}}$, $\text{Belebele}$) & 0.9247 & 0.9421 & 0.8291 & 0.9478 & {0.9588} & 0.8364 & {0.8404} & {0.9732} & 0.8425 & 0.8994 \\
& $\rho$ ($\mu_{\text{Max}}$, $\text{Belebele}$) & {0.9623} & {0.9676} & {0.9211}	& 0.9392 & 0.9326 & 0.8362 & 0.7649 & 0.9448 & 0.9198 & \underline{0.9098}\\
& $\rho$ ($\mu_{\text{Mean}}$, $\text{m-MMLU}$) & {0.9342} & {0.9697} & 0.9362 & {0.9689} & 0.9647 & 0.9223 & {0.9406} & {0.9857} & 0.9393 & \underline{0.9513} \\
& $\rho$ ($\mu_{\text{Max}}$, $\text{m-MMLU}$) & 0.9060 & 0.9596 & 0.8946 & 0.9003 & 0.8892 & 0.9386 & 0.8936 & 0.9311 & 0.9565 & 0.9188	\\
& $\rho$ ($\mu_{\text{Mean}}$, $\text{m-ARC}$) & {0.9741} & {0.9706} & 0.9374 & 0.9515 & 0.9562 & 0.9052 & {0.9268} & {0.9693} & {0.8630} & \textbf{0.9393} \\
& $\rho$ ($\mu_{\text{Max}}$, $\text{m-ARC}$) & 0.9187 & 0.9499 & 0.8736 & 0.8582	& 0.8663 & 0.9297 & 0.8439 & 0.9001 & 0.8298 & 0.8856\\
\cmidrule{2-12}
\multirow{12}{*}{\rotatebox{90}{\phantom{FLORES}}}
\multirow{6}{*}{\rotatebox{90}{\scalebox{.65}{last token}}}
& $\rho$ ($\mu_{\text{Mean}}$, $\text{Belebele}$) & 0.8997 & 0.9326 & 0.8491 & {0.9494} & 0.9581 & {0.9141} & 0.8340 & 0.9679 & {0.9467} & \textbf{0.9168}\\
& $\rho$ ($\mu_{\text{Max}}$, $\text{Belebele}$) & 0.9225 & 0.9309 & 0.9127 & 0.9244 & 0.9123 & 0.9125	& 0.7693 & 0.9460 & 0.9218 & 0.9058 \\
& $\rho$ ($\mu_{\text{Mean}}$, $\text{m-MMLU}$) & 0.9086 & 0.9637 & {0.9370} & 0.9687 & {0.9690} & {0.9771} & 0.9301 & 0.9659 & {0.9700} & \textbf{0.9545} \\
& $\rho$ ($\mu_{\text{Max}}$, $\text{m-MMLU}$) & 0.8448 & 0.9297 & 0.8645 & 0.9224 &  0.9177 & 0.9699 & 0.8902 & 0.9161 & 0.9649 & 0.9134	\\
& $\rho$ ($\mu_{\text{Mean}}$, $\text{m-ARC}$) & 0.9190 & 0.9541 & {0.9524} & {0.9536} & {0.9617} & 0.9390 & 0.9146 & 0.9451 & 0.7356 & \underline{0.9195} \\
& $\rho$ ($\mu_{\text{Max}}$, $\text{m-ARC}$) & 0.8569 & 0.9147 & 0.9005 & 0.8944 & 0.8879 & {0.9464} & 0.8263 & 0.8859 & 0.7037 & 0.8685 \\
\midrule
\midrule
\multirow{12}{*}{\rotatebox{90}{Bible}}
\multirow{6}{*}{\rotatebox{90}{\scalebox{.65}{weighted average}}}
& $\rho$ ($\mu_{\text{Mean}}$, $\text{Belebele}$) & 0.8360 & 0.8530 & 0.7909 & 0.8781 & 0.8974 & 0.8982 & {0.8404} & 0.9118 & 0.7410 & \underline{0.8496} \\
& $\rho$ ($\mu_{\text{Max}}$, $\text{Belebele}$) & 0.{8863} & {0.9001} & {0.8851} & {0.9242} & {0.9302} & 0.8926	& 0.8230 & {0.9337} & 0.7549 & \textbf{0.8811} \\
& $\rho$ ($\mu_{\text{Mean}}$, $\text{m-MMLU}$) & {0.8051} & {0.8886} & 0.8958 & {0.9096} & {0.8964} & {0.9252} & {0.9159} & 0.9093 & 0.7944 & \textbf{0.8823}\\
& $\rho$ ($\mu_{\text{Max}}$, $\text{m-MMLU}$) & 0.5501 & 0.8831	& 0.7748 & 0.8683 & 0.8364 & 0.9180 & 0.9085 & {0.9107} & 0.7388 & \underline{0.8210}\\
& $\rho$ ($\mu_{\text{Mean}}$, $\text{m-ARC}$) & {0.8505} & {0.8998} & 0.9188 & {0.9267} & {0.9116} & {0.8940} & {0.9208}	& {0.9317} & {0.8623} & \textbf{0.9018}\\
& $\rho$ ($\mu_{\text{Max}}$, $\text{m-ARC}$) & 0.6070 & 0.8803 & 0.8030 & 0.8769	 & 0.8552 & 0.8684 & 0.8879 & 0.9178 & 0.8220 & \underline{0.8354}\\
\cmidrule{2-12}
\multirow{12}{*}{\rotatebox{90}{\phantom{Bible}}}
\multirow{6}{*}{\rotatebox{90}{\scalebox{.65}{last token}}}
& $\rho$ ($\mu_{\text{Mean}}$, $\text{Belebele}$) & 0.7656 & 0.8005 & 0.5944 & 0.7934 & 0.8396 & 0.9046	& 0.8299 & 0.9177 & {0.8866} & 0.8147 \\
& $\rho$ ($\mu_{\text{Max}}$, $\text{Belebele}$) & 0.7844 & 0.8299 & 0.5264 & 0.8000 & 0.8100 & {0.9047} & 0.8048 & 0.9235 & 0.8796 & 0.8070 \\
& $\rho$ ($\mu_{\text{Mean}}$, $\text{m-MMLU}$) &  0.7194 & 0.7646 & 0.6472 & 0.6068 & 0.6516 & 0.8827 & 0.8692 & 0.8672	& {0.8060} & 0.7572 \\
& $\rho$ ($\mu_{\text{Max}}$, $\text{m-MMLU}$) & 0.7075 & 0.6886 & 0.5037** & 0.5228**	& 0.4461** & 0.9079 & 0.8576 & 0.8643	& 0.7994 & 0.6998 \\
& $\rho$ ($\mu_{\text{Mean}}$, $\text{m-ARC}$) & 0.7411 & 0.7754 & 0.6592 & 0.5976 & 0.6494 & 0.8537 & 0.8537 & 0.8927 & 0.6997 & 0.7469 \\
& $\rho$ ($\mu_{\text{Max}}$, $\text{m-ARC}$) & 0.7293 & 0.7000 & 0.5190** & 0.5335** & 0.4853** & 0.8494 & 0.8309 & 0.8624 & 0.6867 & 0.6885 \\
\bottomrule
\end{tabular}
}
\caption{Pearson correlation of \genericname using FLORES and Bible data across three tasks. $\rho$ ($\mu_{\text{Pooling}}$, $\text{Task}$) is the correlation of \genericname for the corresponding pooling strategy and benchmark. In all settings except **, the p-value is \( p < 0.001 \). The best average correlations for each task are in \textbf{bold}, and the second bests are \underline{underlined}.}
\tablabel{pearson}
\end{table*}

\subsection{Detailed Results}

We show the detailed per model results of \tabref{pearson_avg} in \tabref{pearson}.

\subsection{\genericname for FLORES-200}\applabel{mexa-whole-flores}

\setlength{\fboxsep}{0pt}

We compute \genericname with weighted average embedding and max pooling for the FLORES parallel data for 203 language labels, multiplied by the performance of Belebele for each model in English. We show the results in \tabref{whole-flores}, and color the cells based on 0.2 intervals from green (well-covered) to red (not covered): 
\colorbox{green!30}{(1.0-0.8)},
\colorbox{limegreen!30}{(0.8-0.6)}, \colorbox{yellow!30}{(0.6-0.4)},
\colorbox{orange!30}{(0.4-0.2)},
\colorbox{red!30}{(0.2-0)}. 
Note that although FLORES is a high-quality, human-translated dataset, we addressed two major issues before proceeding, as noted by \citet{kargaran-etal-2023-glotlid}. First, the data labeled as Cantonese (Yue Chinese) is not actually Cantonese, so we removed it. Second, the code for Central Atlas Tamazight (tzm), which actually refers to Standard Moroccan Tamazight (zgh), was renamed accordingly.
As Belebele is relatively an easy task since the models get good scores in English, and we are using max pooling, this gives a high estimate of the coverage the LLMs have. If the score for a language is not very high, it likely indicates that for more challenging tasks, it will remain low.
In \tabref{whole-flores}, we can clearly see that Llama 3.1-70B and Gemma 2-9B show a higher level of multilinguality than other models.

\onecolumn

{\small\tabcolsep=2.1pt  
\begin{longtable}{ >{\bfseries\footnotesize}ccccccccccc}
\toprule
&  Gemma 2 & Gemma 1 & Llama 3.1 & Llama 3.1 & Llama 3 & Llama 2 & Llama 1 & Mistral & OLMo & AVG \\
& 9B & 7B & 70B & 8B & 8B & 7B & 7B & 7B & 7B \\
\midrule
\endfirsthead

\toprule
&  Gemma 2 & Gemma 1 & Llama 3.1 & Llama 3.1 & Llama 3 & Llama 2 & Llama 1 & Mistral & OLMo & AVG \\
& 9B & 7B & 70B & 8B & 8B & 7B & 7B & 7B & 7B \\
\midrule
\endhead
\midrule
\multicolumn{11}{r}{{Continued on next page}} \\
\midrule
\endfoot

\bottomrule
\endlastfoot
eng\_Latn &      \cellcolor{green!30}0.92 &      \cellcolor{green!30}0.85 &      \cellcolor{green!30}0.95 &      \cellcolor{green!30}0.88 &      \cellcolor{green!30}0.87 &  \cellcolor{yellow!30}0.48 &  \cellcolor{yellow!30}0.42 &      \cellcolor{green!30}0.84 &  \cellcolor{limegreen!30}0.77 &  \cellcolor{limegreen!30}0.77 \\
fra\_Latn &      \cellcolor{green!30}0.92 &      \cellcolor{green!30}0.84 &      \cellcolor{green!30}0.94 &      \cellcolor{green!30}0.88 &      \cellcolor{green!30}0.87 &  \cellcolor{orange!30}0.37 &  \cellcolor{yellow!30}0.41 &      \cellcolor{green!30}0.84 &  \cellcolor{limegreen!30}0.70 &  \cellcolor{limegreen!30}0.75 \\
por\_Latn &      \cellcolor{green!30}0.92 &      \cellcolor{green!30}0.84 &      \cellcolor{green!30}0.94 &      \cellcolor{green!30}0.88 &      \cellcolor{green!30}0.87 &  \cellcolor{yellow!30}0.41 &  \cellcolor{yellow!30}0.41 &      \cellcolor{green!30}0.84 &  \cellcolor{limegreen!30}0.63 &  \cellcolor{limegreen!30}0.75 \\
deu\_Latn &      \cellcolor{green!30}0.92 &      \cellcolor{green!30}0.84 &      \cellcolor{green!30}0.94 &      \cellcolor{green!30}0.88 &      \cellcolor{green!30}0.87 &  \cellcolor{orange!30}0.35 &  \cellcolor{yellow!30}0.42 &      \cellcolor{green!30}0.84 &  \cellcolor{limegreen!30}0.65 &  \cellcolor{limegreen!30}0.74 \\
spa\_Latn &      \cellcolor{green!30}0.92 &      \cellcolor{green!30}0.83 &      \cellcolor{green!30}0.95 &      \cellcolor{green!30}0.88 &      \cellcolor{green!30}0.87 &  \cellcolor{orange!30}0.37 &  \cellcolor{yellow!30}0.42 &      \cellcolor{green!30}0.84 &     \cellcolor{yellow!30}0.56 &  \cellcolor{limegreen!30}0.74 \\
ita\_Latn &      \cellcolor{green!30}0.92 &      \cellcolor{green!30}0.83 &      \cellcolor{green!30}0.92 &      \cellcolor{green!30}0.88 &      \cellcolor{green!30}0.87 &  \cellcolor{orange!30}0.35 &  \cellcolor{yellow!30}0.42 &      \cellcolor{green!30}0.84 &     \cellcolor{yellow!30}0.56 &  \cellcolor{limegreen!30}0.73 \\
\end{longtable}
\twocolumn

\end{document}